\documentclass[acmtog]{acmart}
\AtBeginDocument{%
  }

\copyrightyear{2026}
\acmYear{2026}
\setcopyright{cc}
\setcctype{by}
\acmConference[SA Conference Papers '26]{SIGGRAPH Asia 2026 Conference Papers}{December 01--04, 2026}{Kuala Lumpur, Malaysia}
\acmBooktitle{SIGGRAPH Asia 2026 Conference Papers (SA Conference Papers '26), December 01--04, 2026, Kuala Lumpur, Malaysia}
\acmDOI{10.1145/3829340.3842158}
\acmISBN{979-8-4007-2842-6/2026/12}

\acmSubmissionID{1060}

\def\method{SpatialCrafter}
\def\titlemethod{SpatialCrafter}

\usepackage{amsmath}

\usepackage{booktabs} 
\usepackage{makecell}
\usepackage{xcolor}

\usepackage{lipsum}
\usepackage{tabularx}
\usepackage{multirow}
\usepackage{algpseudocode}

\newcommand{\Isrc}{\mathbf{I}_{0}}     
\newcommand{\Traj}{\mathcal{T}}          
\newcommand{\Pose}{\mathbf{P}}           
\newcommand{\Proxy}{\mathcal{X}}         
\newcommand{\Vgen}{\hat{\mathbf{V}}}     
\newcommand{\Igen}{\hat{\mathbf{I}}}     
\newcommand{\Dgen}{\hat{\mathbf{D}}}     

\newcommand{\Vcond}{\mathcal{V}_{\text{cond}}}   
\newcommand{\Vclean}{\mathcal{V}_0}              
\newcommand{\Vnoisy}{\mathcal{V}_t}              
\newcommand{\Oscene}{\mathcal{X}_{\text{scene}}} 

\newcommand{\PaSS}{PaSS-Flow}     

\newcommand{\Frst}[1]{\textcolor{red}{\textbf{#1}}}
\newcommand{\Scnd}[1]{\textcolor{blue}{\textbf{#1}}}

\newcommand{\tref}[1]{Tab.~\ref{#1}}

\newcommand{\fref}[1]{Fig.~\ref{#1}}
\newcommand{\Fref}[1]{Figure~\ref{#1}}
\newcommand{\sref}[1]{Sec.~\ref{#1}}

\begin{document}

\title{\titlemethod: Single-Image World Modeling with Generative 3D Proxies}

\author{Chuan Fang}
\affiliation{%
  \institution{Hong Kong University of Science and Technology}
 \city{Hong Kong}
 \country{Hong Kong}}
 \email{cfangac@connect.ust.hk}

\author{Lingteng Qiu}
\affiliation{%
  \institution{Tongyi Lab, Alibaba Group}
  \city{Hang Zhou}
  \country{China}}
 \email{220019047@link.cuhk.edu.cn}

\author{Yixun Liang}
\affiliation{%
  \institution{Hong Kong University of Science and Technology}
 \city{Hong Kong}
 \country{Hong Kong}}
 \email{yliang982@connect.ust.hk}

\author{Rui Chen}
\affiliation{%
  \institution{Hong Kong University of Science and Technology}
 \city{Hong Kong}
 \country{Hong Kong}}
 \email{riorui@foxmail.com}

\author{Kunming Luo}
\affiliation{%
  \institution{Hong Kong University of Science and Technology}
 \city{Hong Kong}
 \country{Hong Kong}}
 \email{kluoad@connect.ust.hk}

\author{Zhaohua Zheng}
\affiliation{%
  \institution{ManyCore Tech Inc.}
  \city{Hang Zhou}
  \country{China}}
 \email{zzh20170507@gmail.com}

\author{Tongyuan Bai}
\affiliation{%
  \institution{Jilin University}
  \city{Ji Lin}
  \country{China}}
 \email{baity23@mails.jlu.edu.cn}

\author{Feipeng Tian}
\affiliation{%
  \institution{Hong Kong University of Science and Technology}
 \city{Hong Kong}
 \country{Hong Kong}}
 \email{flybirdtian@gmail.com}

\author{Zilong Dong}
\affiliation{%
  \institution{Tongyi Lab, Alibaba Group}
  \city{Hang Zhou}
  \country{China}}
 \email{zjudzl@qq.com}

\author{Zihan Zhou}
\affiliation{%
  \institution{ManyCore Tech Inc.}
  \city{Hang Zhou}
  \country{China}}
 \email{shuer@qunhemail.com}

\author{Ping Tan}
\affiliation{%
  \institution{Hong Kong University of Science and Technology}
 \city{Hong Kong}
 \country{Hong Kong}}
 \email{pingtan@ust.hk}


\renewcommand{\shortauthors}{FANG et al.}

\begin{abstract}
Explorable image-to-scene generation is essential for applications in gaming, robotics, and virtual reality. Existing methods based on video diffusion model (VDM) commonly rely on incomplete conditioning signals such as sparse point clouds or 2D panoramas, leading to stochastic hallucinations, long-term drifts and suboptimal 3D consistency. We present \method, a novel two-stage framework that addresses these issues by introducing a global 3D proxy for high-fidelity image-to-scene generation. Specifically, we decompose the generation process into global proxy generation and appearance refinement. For proxy generation, we propose a Point-anchored Sparse Structure~(PaSS) Flow module that predicts a spatially aligned and geometrically consistent 3D proxy. For appearance refinement, we re-frame the VDM as a Generative Deferred Refiner which synthesizes high-frequency photorealistic details upon proxy-defined scene geometry. To better integrate the proxy with the pre-trained VDM, we introduce Parallel Geometry Injection and Proxy-Aware Corruption training strategies, which improve robustness to proxy artifacts without disrupting the pretrained generative manifold. Furthermore, as no suitable dataset exists for this explorable scene generation task, we construct a new large-scale dataset of 115K scenes. To the best of our knowledge, it is the first hybrid dataset for image-to-scene generation. Extensive experiments on both synthetic and real-world datasets show that \method~ outperforms state-of-the-art methods, mitigates long-term drift, and remains robust and consistent under rapid camera motion and extreme viewpoint changes. 
Our project page: \href{https://fangchuan.github.io/SpatialCrafter/}{fangchuan.github.io/SpatialCrafter/}
\end{abstract}


\begin{teaserfigure}
  \includegraphics[width=\textwidth]{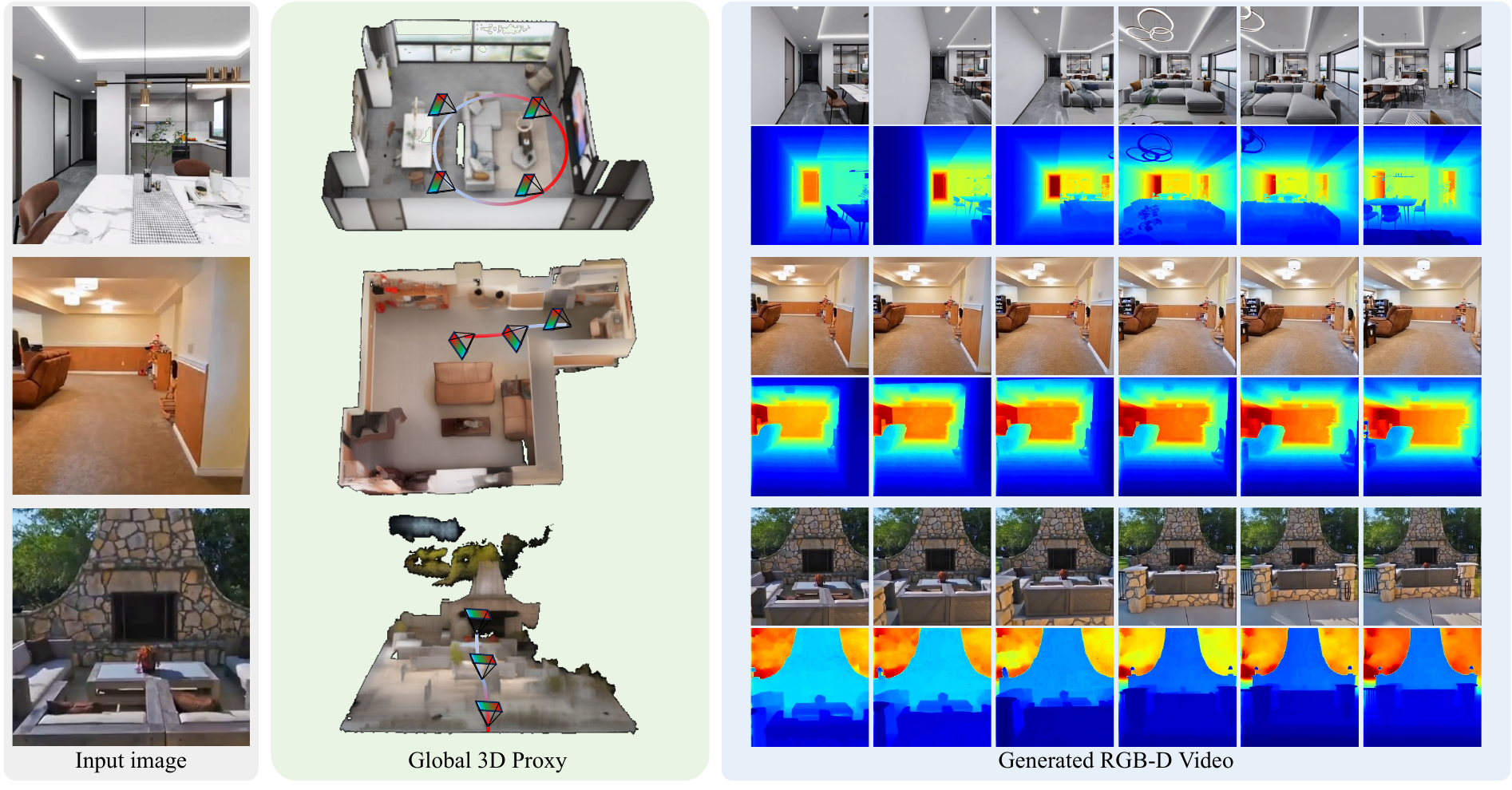}
  \caption{\titlemethod~enables 3D-consistent world modeling from a single image. Given an input image and a user-defined camera trajectory, \titlemethod~first generates a global 3D proxy, then synthesizes a photorealistic RGB-D video that progressively reveals the surrounding environment.}
  \label{fig:teaser}
\end{teaserfigure}

\maketitle

\section{Introduction}
\label{sec:intro}

Building world models is a critical step toward artificial general intelligence (AGI), enabling agents to predict, plan, and reason about complex 3D environments. Central to this vision is the creation of high-quality, diverse 3D environments, which underpin applications ranging from AR/VR content creation and robotic navigation to embodied AI. This ambition has spurred a surge of interest in automated, data-driven 3D scene generation~\cite{zeronvs,SceneCraft,CAT3D,Prometheus,CtrlRoom,Spatialgen}.

Recently, world models built upon video diffusion models (VDMs) have gained significant traction because VDMs offer strong generative priors from massive Internet videos. However, VDMs inherently lack explicit 3D supervision, making it difficult to preserve spatial consistency from a single view. To alleviate this, some works adopt an images-as-memory paradigm, reusing previously generated frames~\cite{DFoT, WorldMem, ContextAsMemory} as visual context. Yet, since these methods remain image-based, they lack the 3D awareness needed for complex camera motions, inevitably causing perspective distortion, occlusion errors, and other geometric inconsistencies.

\begin{figure}[t!]
    \centering
    \includegraphics[width=0.48\textwidth]{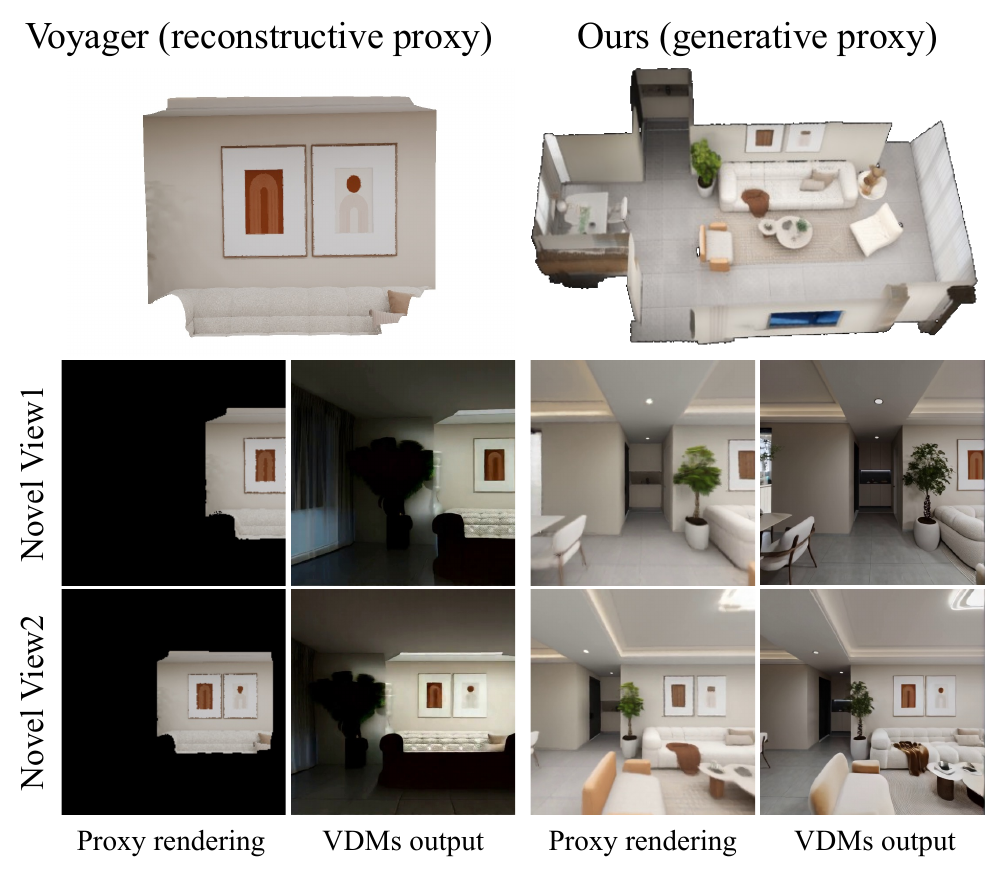}
    \caption{\textbf{Reconstructive vs.\ generative 3D proxies.} Conditioning on an incomplete proxy (partial point clouds), as in Voyager~\cite{Voyager}, leads to severe geometric distortion, inconsistent lighting, and progressive content drift under large camera motion. In contrast, leveraging a global 3D proxy from a separate, native generator, our method yields photometrically and geometrically coherent novel view synthesis (zoom in for details).}
    \label{fig:previous}
\end{figure}

To overcome the limitations of 2D memory, a line of methods~\cite{VMem, Viewcrafter, Gen3c, SPMem, Matrix3D} employs explicit 3D proxies, such as point clouds, as guidance. Given one or more views as input, they first use monocular depth estimation~\cite{MoGev2} or multi-view reconstruction~\cite{DUST3R,VGGT} methods to obtain a point cloud representation of the scene, which is then rendered under new viewpoints to guide the VDM for novel view synthesis. While effective, such proxies are \emph{inherently incomplete} as they are reconstructed only from the input views, as illustrated in \fref{fig:previous}; consequently, VDMs are forced to rely on stochastic hallucination to fill these unseen regions. We refer to this family as \emph{reconstructive} proxies. The \emph{static} variants~\cite{VMem, Viewcrafter, Gen3c, Matrix3D} derive this incomplete proxy once and never update it, while \emph{incremental} variants~\cite{SPMem, Voyager} instead attempt to iteratively update the proxy on the fly, fusing newly generated frames back into the point cloud so that previously hallucinated content anchors later frames. This closes part of the gap, but introduces a new dilemma: the completeness of the proxy now depends on the spatial awareness of the VDM it is meant to enhance. 

In this work, we introduce {\bf \method}, which advances the field of video world models by conditioning VDMs on a new form of 3D proxies that provide dense, continuous, and reliable guidance information. Specially, our framework adopts a two-stage pipeline (see \fref{fig:teaser}). In the first stage, we train a native 3D proxy generator to predict a global scene representation from a single image. With the global proxy, in the second stage, a video diffusion model is re-purposed as a Generative Deferred Refiner that transforms coarse 3D proxy into photorealistic novel views. We compares the effects of our framework (i.e., generative 3D proxies) with prior work (i.e., reconstructive 3D proxies) in \fref{fig:previous}.

Realizing this new framework, however, requires overcoming two key challenges: (1)resolving the spatial misalignment that arises between the input view and the stochastically generated 3D proxy. As noted in prior work ~\cite{reconviagen}, while subtle in object-centric tasks, this misalignment is profoundly amplified in scene-level generation, where even minor discrepancies break cross-view coherence and misguide downstream refinement; (2)generating a global 3D proxy from a single image demands large-scale, high-quality 3D scene data with precise geometric annotations, a resource that remains severely scarce in the community.

\emph{First}, to eliminate spatial misalignment, we introduce the \emph{Point-Anchored Sparse Structure (PaSS) Flow Matching} module. PaSS conditions sparse structure generation on explicit geometric anchors reconstructed from the input view, enforcing strict alignment between the generated 3D proxy and the reference image, thereby providing reliable conditioning for downstream refinement. On the video refinement stage, we propose two complementary mechanisms to convert the coarse 3D proxy into high-quality video. \emph{Parallel Geometry Injection} faithfully preserves scene topology by integrating geometric guidance without disrupting the VDM's pretrained generative manifold. Complementarily, \emph{Proxy-Aware Corruption (PAC)} stochastically perturbs the coarse renderings during training, preventing the refiner from overfitting to proxy-specific artifacts and forcing it to robustly inpaint missing details.

\emph{Second}, to address the data bottleneck, we develop a scalable data engine that combines synthetic indoor scene renderings~\cite{Spatialgen} with real-world videos~\cite{Re10k,DL3DV}. For real footage, we reconstruct consistent 3D geometry via depth and camera pose estimation~\cite{DAv3,vipe} and apply rigorous filtering to discard unreliable reconstructions. This pipeline yields approximately 115K high-fidelity 3D scenes---to the best of our knowledge, the first large-scale dataset tailored for image-to-scene generation, spanning diverse indoor and outdoor environments with precise geometric annotations.


Experiments on synthetic and real-world benchmarks show that \titlemethod~achieves state-of-the-art visual quality and geometric consistency, maintaining robust spatial coherence under extreme camera motion and viewpoint changes where prior methods fail.

\noindent In summary, our contributions are:
\begin{enumerate}
    \item We identify the hallucination problem caused by prior 3D conditioning techniques for video world models and propose \method, a two-stage framework that leverages a global 3D proxy to achieve robust spatial consistency.
    \item We introduce several new techniques for 3D-2D alignment, including the Point-Anchored Sparse Structure Flow Matching (PaSS) module for proxy generation, and the Parallel Geometry Injection and Proxy-Aware Corruption modules for robust video refinement.
    \item We construct the first large-scale, hybrid dataset for image-to-scene generation, comprising 115K scenes with precise geometric annotations across diverse indoor and outdoor environments.
    \item Extensive experiments show that \method~achieves state-of-the-art results, delivering superior spatial consistency even under extreme camera motion and viewpoint changes.
\end{enumerate}

\section{Related Work}
\label{sec:related_work}

\subsection{Image to 3D Scene Generation}
Automated 3D content creation has advanced rapidly through 2D-lifting optimization~\cite{SDS,fantasia3d,EnVision2023luciddreamer,yi2023gaussiandreamer,lin2023magic3d,Iris3D,liu2023syncdreamer,long2023wonder3d,hifi123,shi2023MVDream,Rodin,3DTopia,liang2025unitex,RichDreamer}. These approaches, however, predominantly focus on isolated objects. Recent native 3D generators employing 3D latent diffusion~\cite{shape2vecset,lai2025latticedemocratizehighfidelity3d,Trellis,Chen_2025_Dora,zhao2023michelangelo,feng2025seed3d10imageshighfidelity,sparc3d} have further improved geometric fidelity, but extending these capabilities to complex, full-scene environments remains a formidable challenge.

For scene-level generation, early methods~\cite{zeronvs,SceneCraft,SetTheScene} adapt Score Distillation Sampling (SDS)~\cite{SDS} to optimize 3D representations such as NeRF~\cite{NeRF} or 3DGS~\cite{3DGS}. However, they often suffer from cross-view semantic inconsistency due to the lack of explicit multi-view constraints. To improve efficiency and consistency, a second line of work generates multi-view images or videos using pretrained 2D diffusion models, followed by 3D reconstruction~\cite{CAT3D,ReconX,Spatialgen,Reconfusion} or incremental outpainting~\cite{Text2room,Wonderjourney,Wonderworld}. While these methods avoid per-scene optimization, their underlying generative process relies on 2D RGB priors without explicit 3D reasoning. As a result, they tend to produce overly smooth geometry and low-resolution textures---a ``coarse reality'' that limits their direct deployment in high-fidelity simulations.

\subsection{Controllable Video Generation}
Rather than directly generating the 3D scenes, video diffusion models provides an alternative, more computationally viable path to world modeling. However, due to the inherent limitations of 3D awareness in the video models, especially when dealing with large scenes or camera motions, the introduction of a memory mechanism becomes indispensable. Currently, there are two main categories of algorithms: one implicitly compresses and retrieves historical context (i.e., previous frames), which we refer as the ``2D image memory'' paradigm; and one explicitly adopts 3D representations as memory, which we refer as the ``3D proxy'' paradigm.

\smallskip
\noindent \textbf{2D Image Memory.} A prevalent strategy for maintaining 3D consistency involves compressing and retrieving incremental historical context~\cite{VMem,gu2025long,WorldMem,ContextAsMemory,zhang2026worldstereo}. For instance, Context-as-Memory~\cite{ContextAsMemory} and DFoT~\cite{DFoT} adopt autoregressive 2D history retrieval, treating previously generated frames as an external memory bank to condition future generation processes. Similarly, WorldStereo~\cite{zhang2026worldstereo} leverages a bottom-up geometric memory, incrementally aligning monocular depth estimations into a local point cloud cache in conjunction with a spatial-stereo retrieval module. While these approaches enhance short-term temporal consistency, their incremental nature renders them highly vulnerable to error propagation. Errors in monocular depth estimation or minor misalignments accumulate over time, inevitably resulting in scale drift and severe loop-closure failures during large-scale scene exploration. Furthermore, the maintenance of expanding 2D memory banks and real-time point cloud stitching imposes considerable computational overhead.

\smallskip
\noindent \textbf{3D Proxy.} Another line of research, exemplified by ViewCrafter~\cite{Viewcrafter} and GEN3C~\cite{Gen3c}, takes explicit 3D representations (e.g., point clouds) as memory~\cite{li2025magicworld,liu2025dynamem,geometryforce,zhao2025spatia,zhou2025learning,Voyager}, eliminating the retrieval overhead of 2D memory. This family divides into two branches. \emph{Static} methods~\cite{Viewcrafter,Gen3c,VMem} reconstruct the point cloud once, leaving unseen regions permanently under-constrained, while \emph{incremental} methods~\cite{SPMem,Voyager} instead lift each newly generated frame back into the point cloud, letting hallucinated content anchor later frames. 
However, this introduces a further dilemma: the 3D proxy's construction depends on the spatial awareness of the VDM it is meant to enhance. Our one-shot, generative proxy sidesteps both regimes by predicting the global scene before any video frame is generated, getting rid of neither a single static reconstruction nor the VDM's drifted outputs.

\smallskip
\noindent \textbf{Panorama Proxy.} A related line of work instead expands the input view into a 360\textdegree{} panorama and lifts it to 3D via reconstruction, using the result as visual context for the VDM: Matrix-3D~\cite{Matrix3D} and One2Scene~\cite{one2scene} both complete a panorama from a single image and lift it into a point cloud. As a 2D representation captured from a single optical center, however, a panorama inherits a core limitation of 2D memory: it struggles with occlusion and cannot provide true 3D parallax under large camera translations, since content beyond its original panoramic viewpoint is never modeled in 3D. Our native 3D proxy avoids this bottleneck by generating the global scene directly in 3D, supporting occlusion reasoning under arbitrary trajectories rather than just viewpoint changes around a fixed center.

\smallskip
\noindent \textbf{3D-Aware Diffusion Refinement.} Our video refinement stage is also related to a broader family of works that enhance imperfect 3D reconstructions or renderings with 2D diffusion priors, such as DiFix3D+~\cite{Difix3D}, VideoFrom3D~\cite{VideoFrom3D}, GenFusion~\cite{Genfusion}, and the concurrent Artifixer~\cite{Artifixer}. In particular, DiFix3D+ shares a similar recipe with our Generative Deferred Refiner---render a coarse 3D representation, then refine it with a diffusion model---but differ along three axes. \emph{Interface}: they refine a 3D asset already built from dense, multi-view captures, whereas \method~ constructs its own proxy directly from a single image. \emph{Temporal consistency}: they refine each rendered view independently, whereas our refiner conditions on the full RGB-D video jointly through a video DiT, yielding temporally coherent trajectories rather than independently touched-up frames. \emph{Capability}: their output is bounded by the completeness of the input reconstruction, while \method~ can synthesize and extrapolate previously unseen regions with 3D and temporal consistency.

\begin{figure*}[t]
    \centering
    \includegraphics[width=\linewidth]{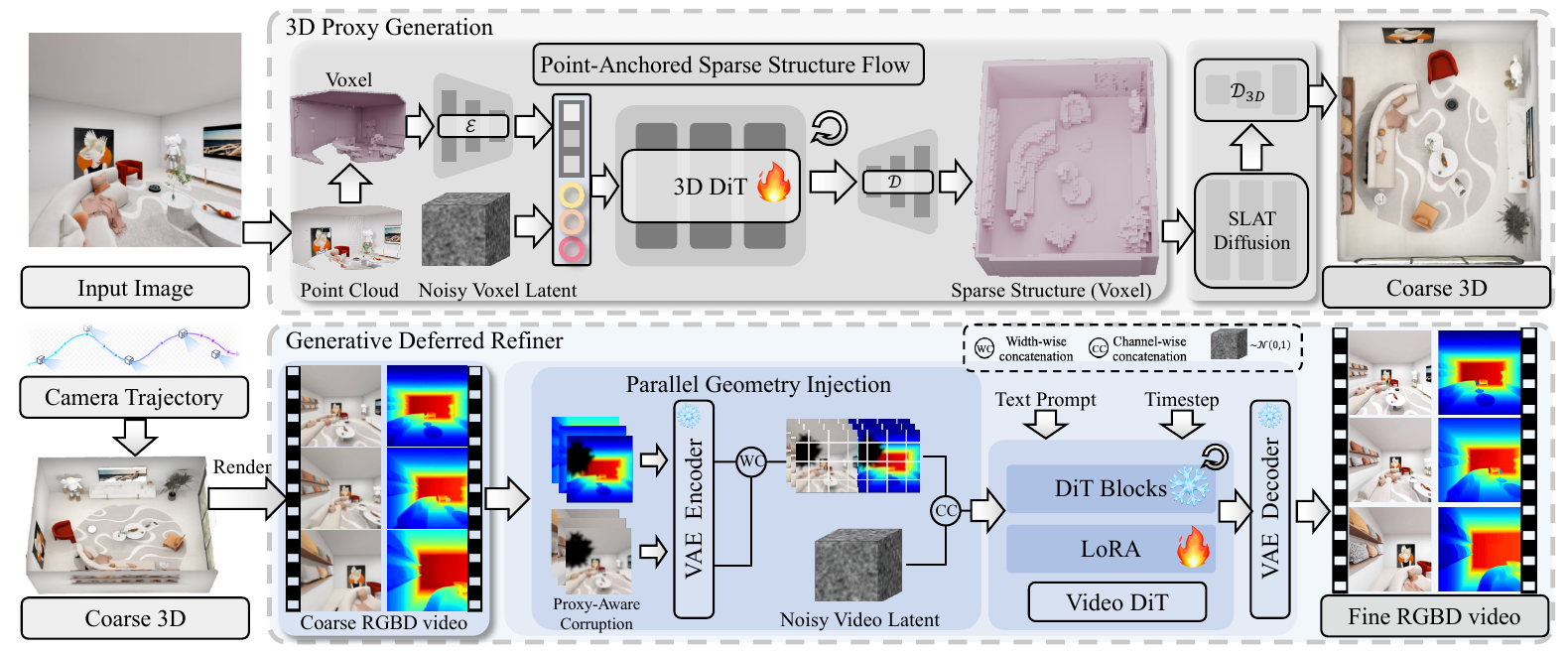}
    \caption{\textbf{Overview of \method}. Given a single image and a camera trajectory, we first construct a global 3D proxy using a native 3D generator with Point-Anchored Sparse Structure (PaSS) Flow Matching. This proxy then serves as a reliable coarse 3D prior for the Generative Deferred Refiner, which transforms it into photorealistic RGB-D video sequences via Parallel Geometry Injection and Proxy-Aware Corruption.} 
    \label{fig:pipeline}
\end{figure*}

\section{Preliminaries}
\label{sec:preliminary}

\noindent \textbf{Latent 3D Diffusion Models.} Our global 3D proxy generation is build upon TRELLIS~\cite{Trellis}, a two-stage framework based on the Structured LATent (SLAT) representation that integrates sparse 3D geometry with multi-view visual features. The coarse stage generates a sparse voxel field $\mathcal{P} = \{\mathbf{p}_i\}_{i=1}^K$, while the refinement stage predicts per-voxel latent features $\mathbf{x}_i$, yielding the complete SLAT $\mathcal{S} = \{(\mathbf{p}_i, \mathbf{x}_i)\}_{i=1}^K$. A 3D Gaussian Splatting~\cite{3DGS} decoder $\mathcal{D}_{\text{3D}}$ then decodes $\mathcal{S}$ into the 3D proxy $\Proxy = \mathcal{D}_{\text{3D}}(\mathcal{S})$.

Both stages use Rectified Flow Transformers~\cite{rectifiedflow} conditioned on DINO embeddings $\mathbf{c}_{\text{img}}$, trained via conditional flow matching (CFM)~\cite{flowmatching} to transport Gaussian noise $\boldsymbol{\epsilon}_{\mathcal{S}} \sim \mathcal{N}(\mathbf{0}, \mathbf{I})$ to the target $\mathcal{S}_0$:
\begin{equation}
    \mathcal{L}_{\text{CFM-3D}}(\phi) = \mathbb{E}_{t, \mathcal{S}_0, \boldsymbol{\epsilon}_{\mathcal{S}}} \left[ \left\| v_\phi(\mathcal{S}_t, t, \mathbf{c}_{\text{img}}) - (\boldsymbol{\epsilon}_{\mathcal{S}} - \mathcal{S}_0) \right\|_2^2 \right],
    \label{eq:trellis_cfm}
\end{equation}
where $\mathcal{S}_t = (1-t)\mathcal{S}_0 + t\boldsymbol{\epsilon}_{\mathcal{S}}$ and $t \in [0, 1]$.

\smallskip
\noindent \textbf{Latent Video Diffusion Models.} Latent video diffusion models operate in a compressed latent space to alleviate the computational burden of high-dimensional video generation. A causal VAE encoder $\mathcal{E}_{\text{vid}}$ maps a video $\mathbf{V} \in \mathbb{R}^{M \times 3 \times H \times W}$ to a compact spatio-temporal latent $\mathbf{z} = \mathcal{E}_{\text{vid}}(\mathbf{V})$. In Wan~2.1~\cite{WanVideo}, the first frame is encoded independently; subsequent frames undergo $4\times$ temporal and $8\times$ spatial downsampling, yielding a 16-channel latent. The video is reconstructed via $\Vgen = \mathcal{D}_{\text{vid}}(\mathbf{z})$.

Video generation likewise follows the CFM framework. A DiT network $v_\theta$, conditioned on text guidance $\mathbf{c}_{\text{txt}}$, predicts the vector field from noise $\boldsymbol{\epsilon}_z \sim \mathcal{N}(\mathbf{0}, \mathbf{I})$ to $\mathbf{z}_0$:
\begin{equation}
    \mathcal{L}_{\text{CFM-Vid}}(\theta) = \mathbb{E}_{t, \mathbf{z}_0, \boldsymbol{\epsilon}_z} \left[ \left\| v_\theta(\mathbf{z}_t, t, \mathbf{c}_{\text{txt}}) - (\boldsymbol{\epsilon}_z - \mathbf{z}_0) \right\|_2^2 \right],
    \label{eq:wan_cfm}
\end{equation}
where $\mathbf{z}_t = (1-t)\mathbf{z}_0 + t\boldsymbol{\epsilon}_z$.

\section{Methodology}
\label{sec:method}

Given a single reference image $\Isrc \in \mathbb{R}^{3 \times H \times W}$ and an arbitrary camera trajectory $\Traj = \{ \Pose_i \}_{i=0}^{N-1}$ of length $N$, where each camera pose $\Pose_i \in SE(3)$, our goal is to synthesize a temporally and spatially consistent video sequence $\Vgen = \{ \Igen_i, \Dgen_i \}_{i=1}^{N-1}$. Here, each generated frame $\Igen_i \in \mathbb{R}^{3 \times H \times W}$ and $\Dgen_i \in \mathbb{R}^{1 \times H \times W}$ denote the predicted RGB image and depth map, respectively, conditioned on the corresponding camera pose $\Pose_i$.

To achieve this, we propose a two-stage framework as illustrated in \fref{fig:pipeline}. In the first stage, we construct a global 3D proxy $\Oscene$ from the input image $\Isrc$ using a native 3D generator~(\sref{sec:3d_proxy_generation}).
In the second stage, a Generative Deferred Refiner synthesizes the target video sequence by refining renderings of $\Oscene$ along the camera trajectory~(\sref{sec:generative_deferred_refiner})
As training both stages requires large-scale, high-quality 3D scene data with precise geometric annotations, we develop a scalable data engine to construct a hybrid dataset from synthetic and real-world sources, detailed in \sref{sec:dataset_construct}.

\subsection{Global 3D Proxy Generator}
\label{sec:3d_proxy_generation}

The primary goal of the first stage is to construct a global 3D scene $\Oscene$ from a single image $\Isrc$, which serves as a complete and reliable explicit spatial prior for the downstream refiner.

\begin{figure}[t] 
    \centering
    \includegraphics[width=0.48\textwidth]{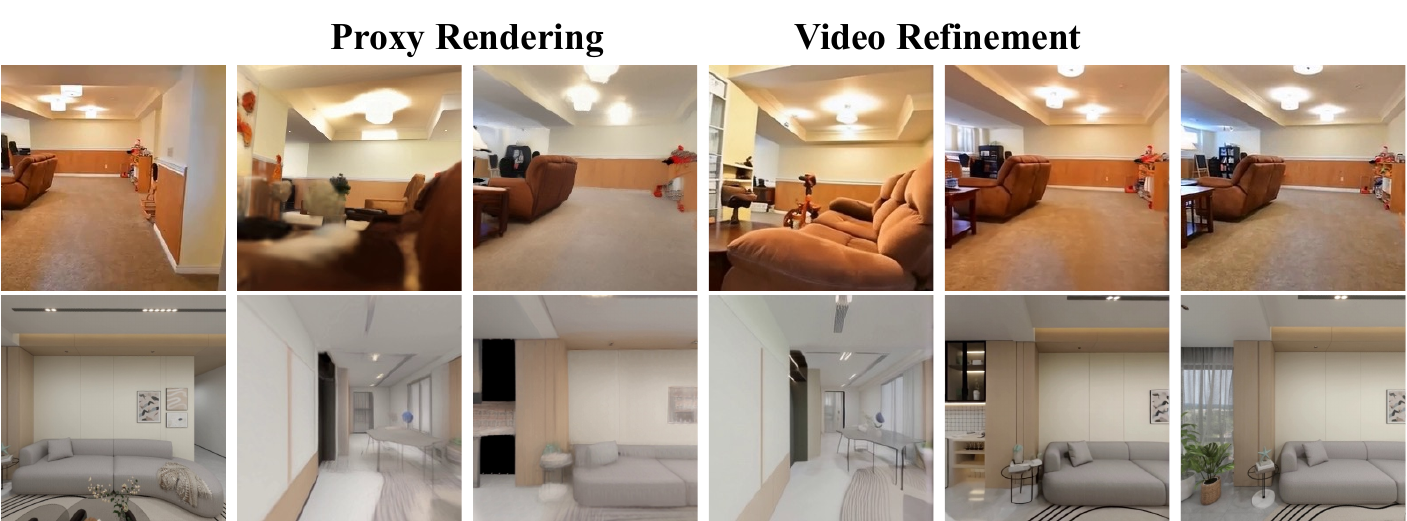}
    \makebox[0.08\textwidth][c]{\footnotesize Input}%
    \makebox[0.08\textwidth][c]{\footnotesize w/o PaSS}%
    \makebox[0.08\textwidth][c]{\footnotesize w/ PaSS}%
    \makebox[0.08\textwidth][c]{\footnotesize w/o PaSS}%
    \makebox[0.08\textwidth][c]{\footnotesize w/ PaSS}%
    \makebox[0.08\textwidth][c]{\footnotesize Ground Truth} 
    \caption{\textbf{Effects of \PaSS~in novel view synthesis.} Without \PaSS, the spatial misalignment between the input view and the generated 3D proxy results in severely degraded proxy renderings, which significantly misguide the video diffusion model and distort the refined novel views. By explicitly conditioning on point anchors, \PaSS~yields structurally coherent coarse renderings that serve as reliable conditioning signals for second-stage refinement. Please zoom in for a better view.}
    \label{fig:ablation_for_pass}
\end{figure}

While existing diffusion-based 3D generators (e.g., TRELLIS~\cite{Trellis}) achieve promising results for object-level generation, their outputs can exhibit spatial misalignment~\cite{reconviagen} with the reference image due to the lack of explicit supervision for view-specific coordinate systems. This misalignment is substantially magnified in scene-level generation. When such a spatially misaligned geometric proxy is employed as the conditioning input for a downstream video refiner, the resulting novel views deviate significantly from the ground truth, as illustrated in \fref{fig:ablation_for_pass}. To address this issue, we introduce the Point-anchored Sparse Structure Flow (PaSS) module.

\smallskip
\noindent \textbf{Point-anchored Sparse Structure Generation.} Our core insight lies in leveraging sparse structural points as explicit geometric anchors, which inherently inject the missing spatial alignment into the view-specific coordinate system, reformulating unconstrained 3D scene generation into a conditioned completion problem.

Specifically, we first unproject the input image $\Isrc$ and its corresponding predicted depth map to construct a partial 3D point cloud. These unprojected points are subsequently transformed into the target coordinate frame using the known pose $\Pose_0$ and voxelized, yielding a set of conditional sparse voxels $\Vcond = \{ \mathbf{p}_{i}^{\text{cond}}\}_{i=1}^{K_\text{ref}}$, where $K_\text{ref}$ denotes the number of active voxels capturing the visible geometry from the input view. Serving as structural anchors for our PaSS module, this explicit geometric prior $\Vcond$ allows us to effectively recast image-conditioned sparse structure generation into a structurally-guided sparse voxel completion task. Under this formulation, the flow-matching model is constrained to synthesize the remaining scene structure such that it is geometrically consistent and perfectly aligned with $\Isrc$.  Consequently, PaSS is optimized via the following objective:
\begin{equation}
    \mathcal{L}_{\text{PaSS}}(\phi_{\text{SS}}) = \mathbb{E}_{t, \Vclean, \boldsymbol{\epsilon}_{\mathcal{V}}} \left[ \left\| v_{\phi_{\text{SS}}}(\mathbf{CAT}(\Vnoisy,\Vcond), t, \Isrc) - (\boldsymbol{\epsilon}_{\mathcal{V}} - \Vclean) \right\|_2^2 \right],
\end{equation}
where $\Vclean$ is the target sparse structure, $\Vnoisy$ its noisy state at timestep $t$, $\boldsymbol{\epsilon}_{\mathcal{V}}$ standard Gaussian noise, and $\mathbf{CAT}(\cdot,\cdot)$ channel-wise concatenation at the DiT input.

Once the aligned sparse structure $\Vclean$ is synthesized, we employ the SLAT Flow to generate the latent features across all active voxels. These features are subsequently decoded by the Gaussian Decoder $\mathcal{D}_{\text{3D}}$ to construct 3D proxy $\Oscene$. 

\subsection{Generative Deferred Refiner}
\label{sec:generative_deferred_refiner}

Given the 3D proxy $\Oscene$, we render the scene along the predefined trajectory $\Traj$ using 3DGS~\cite{3DGS}. This process yields a temporally and spatially consistent sequence of coarse RGB and depth renderings:
\begin{equation}
  \mathbf{c}_{\text{geo}} = \bigl\{ \bigl(\mathbf{I}^{\text{cond}}_i,\,
                         \mathbf{D}^{\text{cond}}_i\bigr)
           \bigr\}_{i=0}^{N-1},
\end{equation}
where $\mathbf{I}^{\text{cond}}_i \in \mathbb{R}^{3 \times H \times W}$ and $\mathbf{D}^{\text{cond}}_i \in \mathbb{R}^{1 \times H \times W}$. This multimodal sequence encodes explicit structural and geometric priors that guide the subsequent refinement stage.

While $\mathbf{c}_{\text{geo}}$ provides robust spatial guidance, it falls short of achieving photorealistic quality. We therefore employ a powerful video foundation model conditioned on $\mathbf{c}_{\text{geo}}$ to refine the coarse renderings into high-fidelity outputs. Following recent image-to-video diffusion models~\cite{WanVideo,hunyuanvideo}, we first encode the reference image $\Isrc$ into a latent representation $\mathbf{z}_{\text{ref}}$. By concatenating this latent token-wise with the noisy video latent, we robustly mitigate distribution drift during subsequent frame synthesis, thereby ensuring that the refined output remains strictly faithful to the original input.

To effectively integrate geometric conditioning signals $\mathbf{c}_{\text{geo}}$, we introduce two novel mechanisms: Parallel Geometry Injection and Proxy-Aware Corruption. Together, they reframe the standard video diffusion model into a Generative Deferred Refiner. Operating analogously to two-pass rendering~\cite{realtimerender} in computer graphics, our refiner explicitly ``repaints'' high-fidelity textures and ``repairs'' geometric artifacts of the coarse 3D proxy (\fref{fig:pipeline}), ultimately delivering photorealistic and visually coherent novel views.

\smallskip
\noindent \textbf{Parallel Geometry Injection.}
To incorporate the coarse 3D proxy without disrupting the VDM's pretrained generative manifold, we first encode the coarse RGB and depth sequences into independent latent representations $\mathbf{z}_{\text{rgb}}$ and $\mathbf{z}_{\text{depth}}$ via the VAE encoder $\mathcal{E}_{\text{vid}}$. These latents are subsequently tiled side-by-side along the width dimension to construct a joint geometric latent $\mathbf{z}_{\text{geo}}$.

The input to the video DiT blocks is constructed by first fusing $\mathbf{z}_t$ and $\mathbf{z}_{\text{geo}}$ along the channel dimension, and then appending the reference latent $\mathbf{z}_{\text{ref}}$ along the token dimension. The training objective follows a conditional flow matching formulation:
\begin{equation}
    \mathcal{L}_{\text{Vid}}(\theta) = \mathbb{E}_{t, \mathbf{z}_0, \boldsymbol{\epsilon}_z} \left[ \left\| v_\theta\bigl( [\mathbf{z}_{\text{ref}},\; \mathbf{CAT}(\mathbf{z}_t, \mathbf{z}_{\text{geo}})],\; t,\; \mathbf{c}_{\text{txt}}\bigr) - (\boldsymbol{\epsilon}_z - \mathbf{z}_0) \right\|_2^2 \right],
    \label{eq:generative_refiner}
\end{equation}
where $\mathbf{CAT}(\cdot,\cdot)$ and $[\cdot,\cdot]$ denote channel-wise and token-wise concatenation, respectively, and $\mathbf{c}_{\text{txt}}$ represents the textual condition. We freeze all pre-trained model weights and only optimize a small set of LoRA~\cite{lora} parameters within the DiT blocks. This strategy allows the model to progressively incorporate geometric guidance while preserving its inherent spatio-temporal and photorealistic priors, thereby ensuring geometrically consistent and visually plausible appearance rendering.

\smallskip
\noindent \textbf{Proxy-Aware Corruption.}
A critical challenge in our pipeline is the fidelity gap between ideal training signals and the imperfect proxies generated during inference. These generated proxies often exhibit artifacts such as over-smoothed textures, high-frequency floaters, and unobserved geometry (holes). Training the refiner exclusively on curated data leads to overfitting, severely degrading its robustness against these out-of-distribution artifacts.

To bridge this gap, we propose Proxy-Aware Corruption (PAC), a stochastic perturbation strategy applied to the geometric conditions $\mathbf{c}_{\text{geo}}$ during training. For each conditioning pair $(\mathbf{I}^{\text{cond}}_i, \mathbf{D}^{\text{cond}}_i)$, we randomly apply one of three degradations: adaptive Gaussian blur, block-wise noise, or partial depth erasure. 
Leveraging PAC, the refiner not only learns to autonomously determine where to anchor onto the physical structure and where to synthesize high-frequency textures and sharp boundaries, but also substantially improves the model's robustness in out-of-domain scenarios.

\begin{figure}[t]
    \centering
    \includegraphics[width=\linewidth]{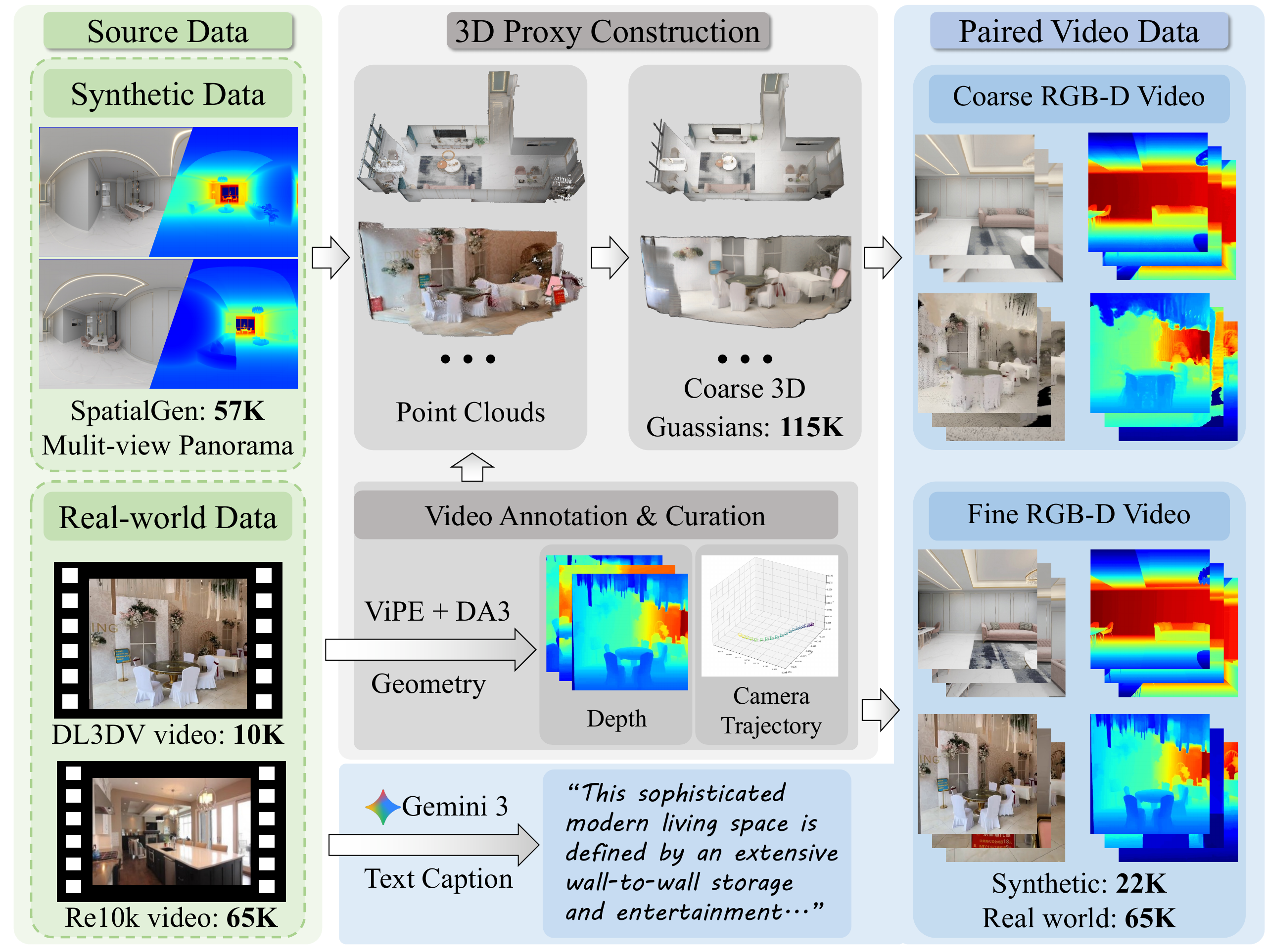}
    \caption{\textbf{Dataset construction pipeline}. We process hybrid data sources by extracting synthetic point clouds and estimating depth for real-world footage, then decode the geometry into coarse 3D Gaussians and render them along camera trajectories to form coarse RGB-D sequences. Pairing these with clean ground-truth videos and text captions yields the large-scale dataset used to train both stages of our framework.}
    \label{fig:data_pipeline}
\end{figure}

\subsection{High-Quality Scene Dataset Construction}
\label{sec:dataset_construct}

Training the 3D proxy generator and the Generative Deferred Refiner presented above demands large-scale, high-quality 3D scene data with precise geometric annotations. However, existing benchmarks~\cite{scannet,Matterport3D,replica,hypersim} are limited in both scale and scene diversity. To overcome this, we introduce a scalable data engine that constructs approximately 115K 3D scenes, seamlessly bridging synthetic indoor renderings and real-world video captures. To the best of our knowledge, this constitutes the first large-scale dataset tailored for single-view 3D scene generation, offering precise geometric annotations and diverse scene categories.

\smallskip
\noindent \textbf{Data Sources.}
We build our dataset with raw data from three main sources, namely SpatialGen~\cite{Spatialgen}, RealEstate10K~\cite{Re10k}, and DL3DV~\cite{DL3DV}. SpatialGen is a large-scale indoor dataset comprising over 4.7M panoramic RGB-D renderings across 57,431 rooms. RealEstate10K and DL3DV are real-world video datasets containing 65,683 and 10,510 videos, respectively, spanning both indoor and outdoor environments.

\smallskip
\noindent \textbf{3D Proxy Construction.}
The first step of 3D proxy construction is to obtain a point cloud of each scene. For SpatialGen, we extract perspective RGB-D images from each panoramic rendering via equi-to-perspective projection~\cite{equilib}, then unproject the RGB-D data using the associated camera poses to reconstruct dense point clouds. For each video in RealEstate10K and DL3DV, we estimate dense depth maps using ViPE~\cite{vipe} in conjunction with DAv3~\cite{DAv3} to enforce cross-view consistency. Together with the provided camera poses, the depth maps are then used to unproject the RGB frames into point cloud. 

Next, we adopt the data processing pipeline of TRELLIS to generate SLAT representations from these point clouds and multi-view images, which are subsequently decoded into coarse 3D Gaussians via $\mathcal{D}_{\text{3D}}$. Crucially, we pre-compute individual SLAT features for each reference image; these serve as a geometric prior for the proposed PaSS module, enabling global proxy generation.

\smallskip
\noindent \textbf{Paired Video Data Construction.}
The original SpatialGen dataset provides panoramic renderings at $0.5$m intervals, which lack the temporal density required for training a video diffusion model. To address this, we select 2.2K high-quality scenes and re-render them along ten distinct, continuous camera trajectories per scene (see Appendix), yielding 22,421 temporally continuous RGB-D video sequences. Meanwhile, for real-world captures in RealEstate10K and DL3DV, we employ a rigorous protocol to filter out scenes with unreliable reconstructions (see Appendix), yielding a set of clean, high-quality real-world RGB-D sequences. 

Next, using the coarse 3D Gaussians constructed above and the recorded camera trajectories, we render low-quality coarse RGB-D sequences aligned with their clean ground-truth counterparts, producing the exact paired data required to train the video refinement model. 
Further, we use Gemini~3~\cite{gemini} to generate captions for every clean video sequence.
The final aggregated dataset comprises 115,295 pairs for image-to-scene generation and 87,115 pairs for controllable video generation.



\begin{table*}[ht!]
    \centering
    \caption{Quantitative comparison on SpatialGen-Video, RealEstate10K~\cite{Re10k}, and DL3DV~\cite{DL3DV} datasets for single image to video generation. Best and second best results are highlighted in \Frst{red} and \Scnd{blue}, respectively.}
    \label{table:img2video}
    \resizebox{\textwidth}{!}{
    \begin{tabular}{l cccccc ccccc ccccc}
        \toprule
        \multirow{2}{*}{Method} & \multicolumn{6}{c}{SpatialGen-Video~\cite{Spatialgen}} & \multicolumn{5}{c}{Re10K~\cite{Re10k}} & \multicolumn{5}{c}{DL3DV~\cite{DL3DV}} \\
        \cmidrule(lr){2-7} \cmidrule(lr){8-12} \cmidrule(lr){13-17}
        & FVD$\downarrow$ & PSNR$\uparrow$ & SSIM$\uparrow$ & LPIPS$\downarrow$ & RPE$\downarrow$ & RVE (rFID$\downarrow$) 
        & FVD$\downarrow$ & PSNR$\uparrow$ & SSIM$\uparrow$ & LPIPS$\downarrow$ & RPE$\downarrow$  
        & FVD$\downarrow$ & PSNR$\uparrow$ & SSIM$\uparrow$ & LPIPS$\downarrow$ & RPE$\downarrow$ \\
        \midrule
        DFoT~\cite{DFoT} & 771.84 & 12.678 & 0.442 & 0.607 & 0.209 & 84.11 & 277.54 & 15.369 & 0.567 & 0.315 & 0.118 & 691.17 & 10.375 & 0.219 & 0.601 & 0.243 \\
        GeometryForcing~\cite{geometryforce} & 661.71 & 12.722 & 0.483 & 0.541 & 0.197 & 70.44 & \Scnd{190.41} & \Scnd{16.46} & \Scnd{0.627} & \Scnd{0.28} & 0.093 & 643.50 & 10.652 & 0.236 & 0.580 & 0.235 \\
        \midrule
        GEN3C~\cite{Gen3c} & 525.23 & 13.718 & 0.524 & 0.522 & 0.151 & \Frst{26.60} & 199.57 & 15.00 & 0.541 & 0.351 & 0.118 & \Scnd{369.99} & 12.155 & 0.276 & 0.554 & 0.189 \\
        Voyager~\cite{Voyager} & 801.04 & 8.204 & 0.302 & 0.66 & 0.278 & 317.0 & 414.72 & 11.605 & 0.374 & 0.488 & 0.144 & 920.27 & 8.792 & 0.178 & 0.683 & 0.219 \\
        ViewCrafter~\cite{Viewcrafter} & \Scnd{339.04} & \Scnd{14.098} & \Scnd{0.555} & \Scnd{0.471} & 0.125 & 41.51 & 347.29 & 14.95 & 0.613 & 0.363 & 0.119 & 470.05 & \Scnd{13.036} & \Scnd{0.334} & \Scnd{0.522} & \Frst{0.147} \\
        \midrule
        Voyager$^\dagger$ & 363.39 & 11.412 & 0.412 & 0.583 & 0.192 & 250.98 & 401.91 & 13.840 & 0.473 & 0.458 & 0.116 & 708.42 & 10.30 & 0.215 & 0.620 & 0.211 \\
        \midrule
        \method & \Frst{193.54} & \Frst{14.775} & \Frst{0.579} & \Frst{0.416} & \Frst{0.093} & \Scnd{38.69} & \Frst{148.71} & \Frst{17.185} & \Frst{0.659} & \Frst{0.266} & \Frst{0.078} & \Frst{242.30} & \Frst{13.356} & \Frst{0.356} & \Frst{0.453} & \Scnd{0.149} \\
    
        \bottomrule
    \end{tabular}
    }
\end{table*}

\section{Experiments}
\subsection{Experiment Setup} 
\label{sec:exp:detail}
\smallskip
\noindent \textbf{Benchmark Datasets.}
We evaluate \titlemethod~on diverse synthetic and real-world benchmarks. For synthetic evaluation, we construct SpatialGen-Video: 104 scenes from SpatialGen rendered with RGB-D ground truth along circular-loop trajectories. For real-world evaluation, we randomly sample 100 RealEstate10K test scenes and adopt the standard DL3DV test split. All experiments use camera poses and depth maps from our data engine for fair comparison.

\smallskip
\noindent \textbf{Baselines.} We compare \titlemethod~against state-of-the-art methods with two conditioning paradigms: 2D image memory (GeometryForcing~\cite{geometryforce}, DFoT~\cite{DFoT}) and reconstructive 3D proxies (ViewCrafter~\cite{Viewcrafter}, GEN3C~\cite{Gen3c}, Voyager~\cite{Voyager}). Voyager is most closely related to our approach, as it likewise generates RGB-D video as its final output. To further isolate the benefit of our proposed high-quality 3D scene dataset from that of our architecture, we faithfully reproduce Voyager's training pipeline and fine-tune it on our dataset for controllable video generation, denoted as Voyager$^\dagger$.

\smallskip
\noindent \textbf{Metrics.} Visual quality is measured by FVD~\cite{FVD}, PSNR, LPIPS~\cite{LPIPS}, and SSIM~\cite{SSIM} against ground-truth sequences. Geometric consistency is evaluated with Re-Projection Error (RPE) and Revisit Error (RVE)~\cite{geometryforce}. Instead of re-estimating depth and poses via DROID-SLAM~\cite{droidslam}, we compute these metrics directly from ground-truth RGB-D of reference image and camera trajectories (see Appendix), isolating evaluation from auxiliary estimator errors. RVE is evaluated only on SpatialGen-Video, whose closed-loop trajectories guarantee identical first and final frames.

All methods generate 81 frames from a single image and camera trajectory---a setting substantially more demanding than prior protocols---using publicly available checkpoints under identical input conditions.

\smallskip
\noindent \textbf{Implementation Details.}
We implement \titlemethod~in PyTorch with a two-stage progressive training strategy. Stage~1 fine-tunes the 3D proxy generator from TRELLIS-Image~\cite{Trellis}: first on curated RealEstate10K and DL3DV scenes for 10K steps, then on SpatialGen for another 10K steps, using 16 NVIDIA H20 GPUs with a batch size of 128 (20K total steps). Stage~2 trains the Generative Deferred Refiner, built on Wan2.1-Fun-Control~\cite{wanfunctrl}, with a progressive resolution schedule: $256\times256$ for 12K steps, then $512\times512$ for 4K steps (81 frames each), using 32 NVIDIA H20 GPUs with batch size 32 (16K total steps). Both stages use AdamW~\cite{kingma2014adam} with an initial learning rate of $10^{-4}$, decaying by $0.01$ at 90\% of training.

\subsection{Experimental Results}
\label{sec:exp:exp_results}
\smallskip
\noindent \textbf{Quantitative Results.}
As shown in \tref{table:img2video}, \titlemethod~consistently achieves superior performance across all three benchmark datasets. On SpatialGen-Video, our method attains the lowest FVD (193.54) and RPE (0.093), demonstrating effective mitigation of long-term drift and robust 3D consistency, surpassing approaches that rely on reconstructive conditioning. The advantage carries over to real-world scenarios: on RE10K, \titlemethod~significantly leads baselines across all key metrics, including an FVD of 148.71 and PSNR of 17.185. On the challenging DL3DV dataset, our approach maintains dominance with the best image quality scores and competitive geometric accuracy. These results consistently validate the superiority and robustness of the proposed method.

Fine-tuning Voyager on our proposed dataset (Voyager$^\dagger$) consistently improves its FVD and PSNR over the original Voyager across all three benchmarks, confirming the value of our high-quality data. However, Voyager$^\dagger$ still lags far behind \method~in geometric consistency, particularly under extensive camera motion, as reflected by its substantially higher RPE and RVE on SpatialGen-Video and DL3DV. We attribute this gap to the underlying reconstructive proxy: even with higher-quality training data, Voyager's incomplete, incrementally-updated point cloud still fails to constrain the VDM, which continues to hallucinate unseen regions and yields inferior 3D consistency. This indicates that our gains stem not merely from the proposed dataset, but fundamentally from the one-shot generative 3D proxy design. 

\smallskip
\noindent \textbf{Qualitative Results.}
Qualitative comparisons in \fref{fig:image2video_comp} further highlight our advantage under challenging long-range camera trajectories and extreme viewpoint changes. Voyager suffers severe structural collapse, while GEN3C and ViewCrafter exhibit prominent content artifacts and geometric distortions. DFoT and GeometryForcing, although more stable, struggle with intense camera motion and produce nearly static outputs on SpatialGen-Video and DL3DV, resulting in noticeable misalignment. In contrast, our method robustly preserves spatially consistent geometry and visually plausible appearance aligned with the input image.

Overall, these experiments confirm that baselines relying on reconstructive 3D proxies or 2D image memories are highly susceptible to long-term drift and stochastic hallucinations. By leveraging a global, complete 3D proxy, \titlemethod~effectively overcomes these limitations, enabling robust and consistent video generation under demanding conditions.
\subsection{Ablation Study}
\label{sec:exp:ablation}
We conduct ablation studies on the two core components of~\titlemethod: the 3D proxy generator and Generative Deferred Refiner.

\smallskip
\noindent \textbf{Effectiveness of \PaSS.}
We compare two variants of our proxy generator on the SpatialGen-Video test set: one trained without \PaSS, and one with \PaSS. As \Fref{fig:ablation_for_pass} shows, omitting \PaSS causes severe spatial misalignment between the input view and the generated proxy. This produces degraded renderings that misguide the downstream refiner. Conversely, by conditioning on point anchors, \PaSS yields structurally aligned coarse renderings that provide reliable signals for high-fidelity refinement. Quantitatively, the \PaSS-equipped model outperforms the baseline across all metrics (\tref{table:ablation_on_spatialgen}).

\begin{figure*}[p]
    \centering
    \vspace{-4mm}
    \includegraphics[width=\textwidth]{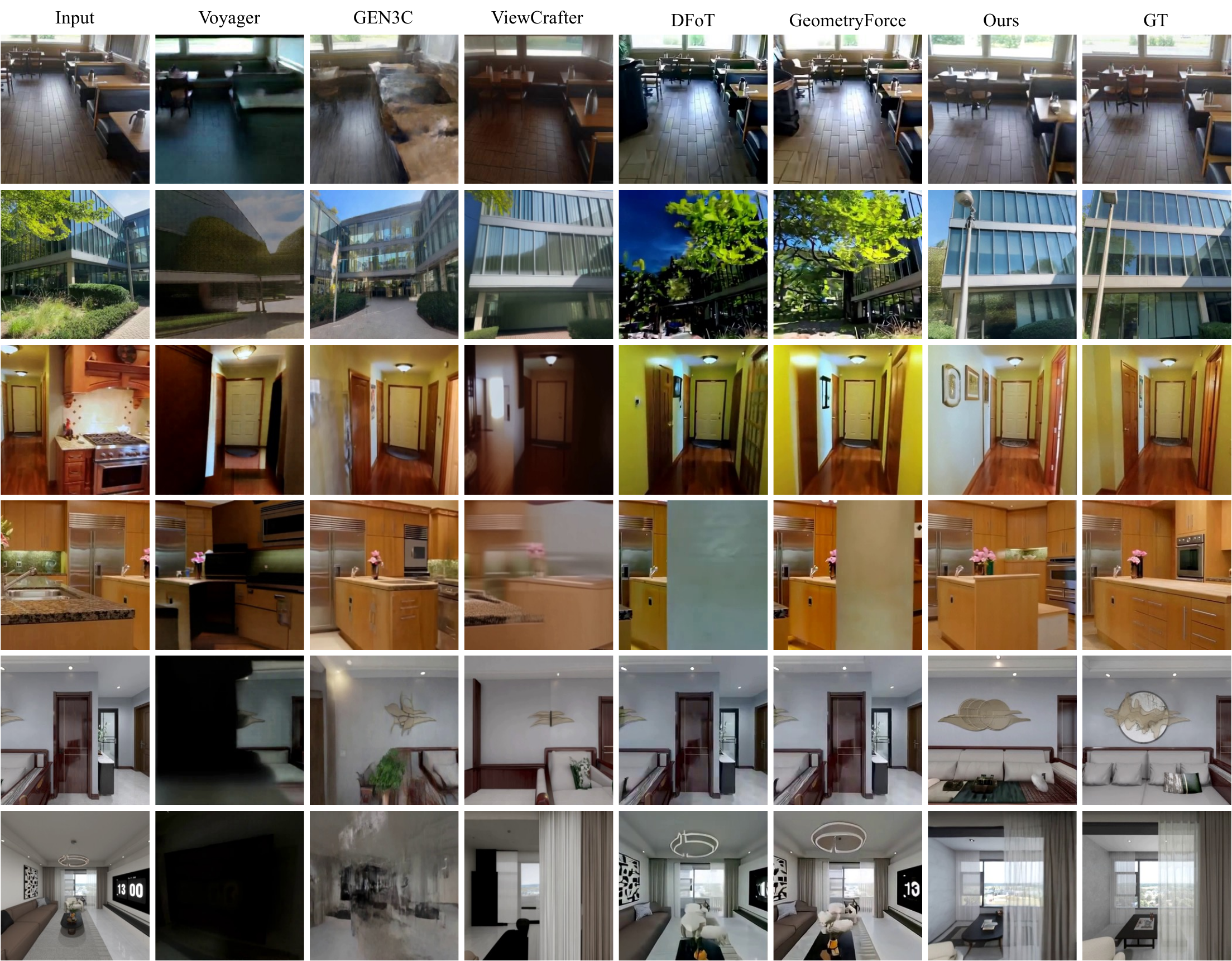}
    \vspace{-5mm}
    \caption{\textbf{Qualitative comparison with SOTA methods.} \titlemethod~significantly surpasses the baseline methods under extreme challenging camera motions, yielding photorealistic and spatially consistent novel view synthesis. Please refer to the supplementary material for more comparison results.}
    \label{fig:image2video_comp}
    \vspace{2mm}
    
    \includegraphics[width=0.47\textwidth]{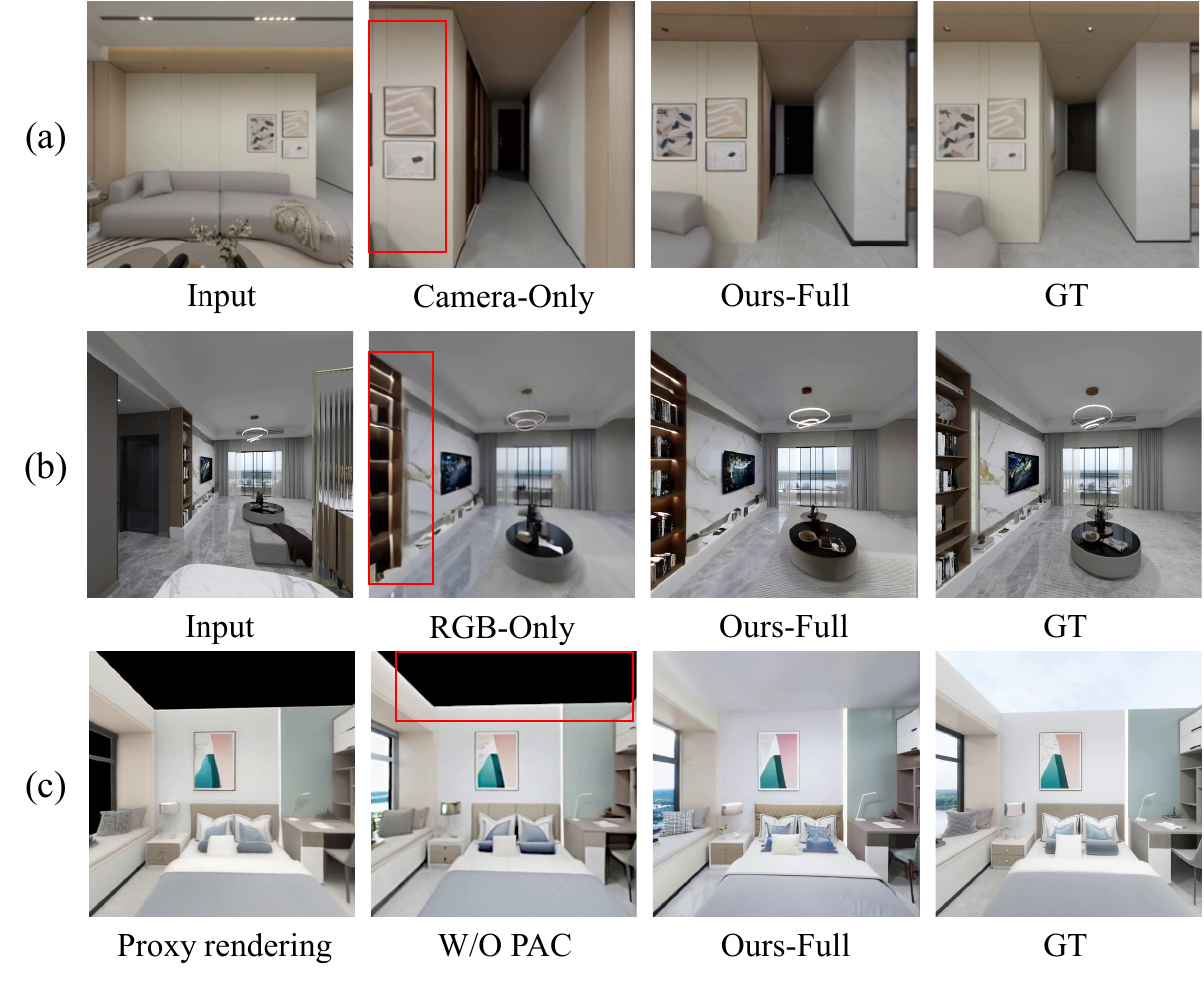}
    \vspace{-4mm}
    \caption{\textbf{Qualitative results on ablation study.} We compare the video models in our four training stages. Our full model achieves the highest quality.}
    \label{fig:ablate_generate_deferred_refiner}
\end{figure*}

\smallskip
\noindent \textbf{Generative Deferred Refiner Configurations.}
To independently assess the design choices within our refinement stage, we use the ground-truth coarse renderings as conditions on the SpatialGen-Video test set and evaluate four configurations: (a)~an alternative baseline that takes camera parameters as conditioning signals rather than coarse proxy renderings, (b)~an RGB-only variant conditioned solely on coarse RGB renderings, (c)~an RGB-D variant conditioned on RGB-D renderings without Proxy-Aware Corruption (PAC), and (d)our full model trained on RGB-D conditions with PAC enabled. As summarized in \tref{table:ablation_on_spatialgen}, conditioning on coarse proxy renderings substantially improves spatial consistency over the camera-only baseline. Adding coarse depth further enhances 3D consistency, and incorporating PAC yields an additional performance gain, producing the most robust and photorealistic novel-view synthesis acr\-oss all metrics. Qualitative comparisons in~\fref{fig:ablate_generate_deferred_refiner} corroborate these findings. As shown in~\fref{fig:ablate_generate_deferred_refiner}(a), the camera-only baseline fails to provide precise spatial control under extreme camera motion, resulting in noticeable drift. \Fref{fig:ablate_generate_deferred_refiner}(b) reveals that the RGB-only variant may generate inconsistent content when the camera moves into poorly observed regions. \Fref{fig:ablate_generate_deferred_refiner}(c) highlights the necessity of PAC: when coarse proxy renderings contain large black areas from unobserved geometry, the RGB-D variant without PAC tends to preserve these artifacts, whereas our full model with PAC robustly inpaints the missing regions, producing spatially coherent and photorealistic outputs.
\begin{table}[ht]
    \centering
    \footnotesize
    \renewcommand{\tabcolsep}{1.5pt}
    \caption{\textbf{Ablation study on SpatialGen-Video.}}
    \label{table:ablation_on_spatialgen}
    \begin{tabularx}{\linewidth}{l *{6}{>{\centering\arraybackslash}X}}
        \toprule
        Method & {FVD} $\downarrow$ & {PSNR} $\uparrow$ & {SSIM} $\uparrow$ & {LPIPS} $\downarrow$ & {RPE} $\downarrow$ & {RVE} $\downarrow$ \\
        \midrule
        Ours~(w/o \PaSS) & 472.07 & 12.386 & 0.492 & 0.577 & 0.169 & 153.47 \\
        Ours~(w/ \PaSS) & \textbf{193.54} & \textbf{14.775} & \textbf{0.579} & \textbf{0.416} & \textbf{0.093} & \textbf{38.69} \\
        \midrule
        Ours~(Camera-only) & 178.47 & 14.50 & 0.565 & 0.435 & 0.107 & 47.81 \\
        Ours~(RGB-only) & 136.27 & 19.47 & 0.681 & 0.287 & 0.066 & 38.69 \\
        Ours~(RGB-D)    & 114.13 & 19.99 & \textbf{0.743} & 0.251 & 0.062 & 32.17 \\
        \midrule
        Ours~(Full)     & \textbf{107.03} & \textbf{20.39} & \textbf{0.743} & \textbf{0.249}   & \textbf{0.057} & \textbf{30.96} \\
        \bottomrule
    \end{tabularx}
\end{table}

\section{Conclusion}
We introduce \titlemethod, a novel two-stage framework for explorable image-to-scene generation. While existing methods are highly susceptible to long-term drift and stochastic hallucinations due to their reliance on incomplete 3D proxies or 2D images as spatial memory, our approach overcomes these limitations by leveraging a global, complete 3D proxy. To tackle the severe scarcity of high-quality training data, we develop a scalable data engine to construct the first large-scale dataset tailored for single-view 3D scene generation. 
Technically, we propose a Point-Anchored Sparse Structure (PaSS) Flow Matching module to enforce geometric alignment, and complement it with Parallel Geometry Injection and Proxy-Aware Corruption for artifact-robust video refinement. Extensive experiments demonstrate that \titlemethod~significantly outperforms existing baselines, delivering geometrically consistent and visually plausible novel view synthesis.

\section*{Acknowledgments}

This work was supported in part by the Key R\&D Program of Zhejiang Province (No.~2026SDXT005) and by HKUST Project No.~24251-090T019. We thank Susu Zheng, Jiakai Li, Feng Chen, Zhaorong Li, Liangbin Hu, and Fuchun Dong, all from Manycore Tech, for their assistance in constructing the synthetic indoor scene dataset. We also thank Chao Xu for rendering additional open-world synthetic data, and Jia Zheng for invaluable suggestions during the preliminary stage of this research.

\begin{figure*}[p]
    \centering
    \includegraphics[width=\textwidth]{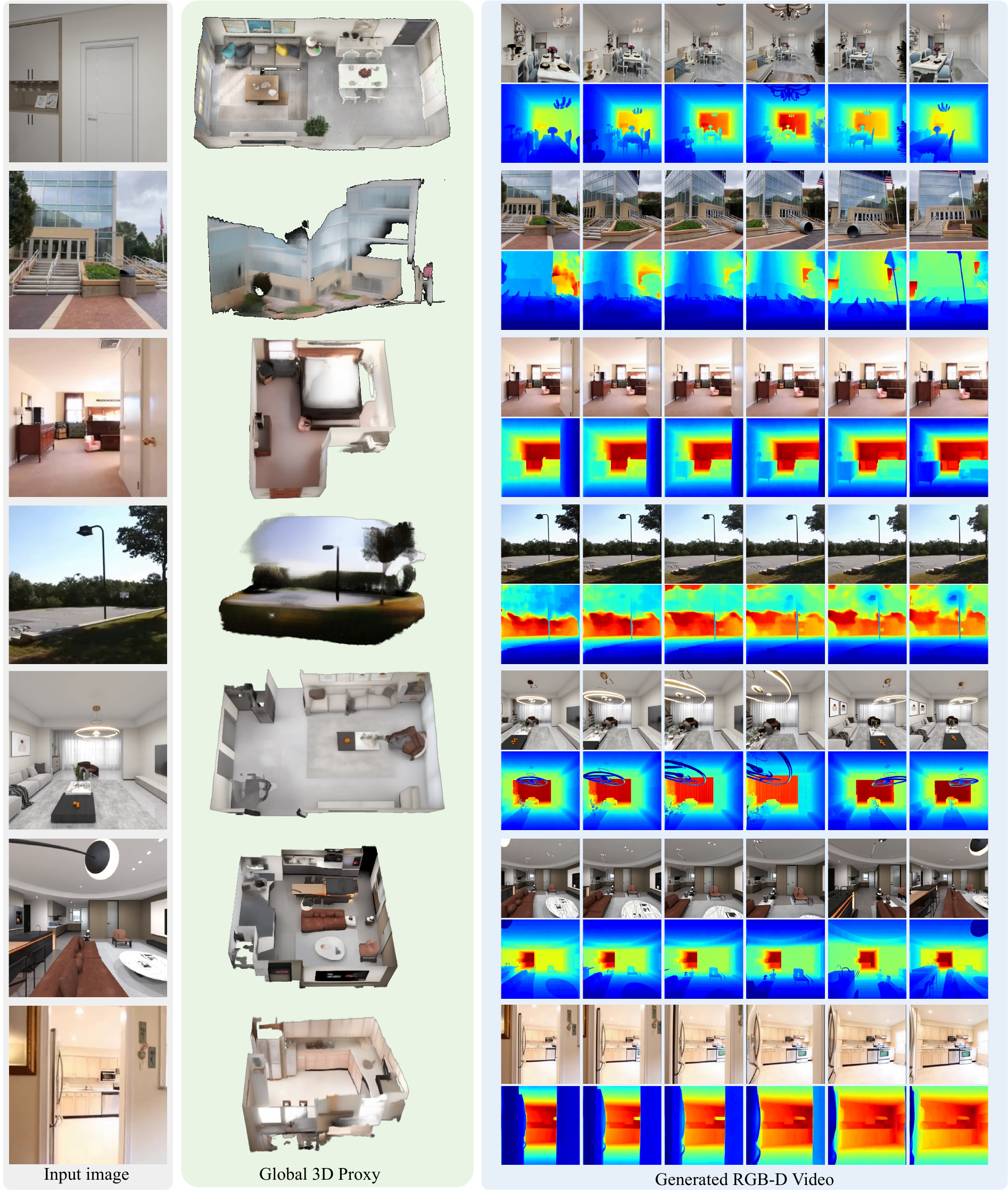}
    \vspace{-6mm}
    \caption{More visual results of \titlemethod.} 
    \label{fig:more_teaser}
\end{figure*}

\bibliographystyle{ACM-Reference-Format}
\bibliography{spatialcrafter}


\end{document}